\documentclass{article} 
\usepackage{iclr2025_conference,times}

\usepackage{amsmath,amsfonts,bm}

\def\eqref#1{equation~\ref{#1}}

\def\1{\bm{1}}

\DeclareMathAlphabet{\mathsfit}{\encodingdefault}{\sfdefault}{m}{sl}
\SetMathAlphabet{\mathsfit}{bold}{\encodingdefault}{\sfdefault}{bx}{n}

\usepackage{hyperref}
\usepackage{url}

\usepackage{multirow}
\usepackage{multicol}
\usepackage{hyperref}
\usepackage{array}
\usepackage{longtable}
\usepackage{pythonhighlight}
\usepackage{booktabs}

\usepackage{amssymb}
\usepackage{mathtools}

\usepackage{svg}
\usepackage{subcaption} 

\iclrfinalcopy

\title{Quantile Regression for  Distributional \\ Reward Models in RLHF}

\author{
Nicolai Dorka \\
~University of Freiburg, Agile Loop ~~~~~  \\
~Correspondence to~\texttt{nicolai.dorka1@gmail.com}
}

\begin{document}

\maketitle

\begin{abstract}
Reinforcement learning from human feedback (RLHF) has become a key method for aligning large language models (LLMs) with human preferences through the use of reward models.
However, traditional reward models typically generate point estimates, which oversimplify the diversity and complexity of human values and preferences.
In this paper, we introduce Quantile Reward Models (QRMs), a novel approach to reward modeling that learns a distribution over rewards instead of a single scalar value. 
Our method uses quantile regression to estimate a full, potentially multimodal  distribution over preferences, providing a more powerful and nuanced representation of preferences.
This distributional approach can better capture the diversity of human values, addresses label noise, and accommodates conflicting preferences by modeling them as distinct modes in the distribution.
Our experimental results show that QRM outperforms comparable traditional point-estimate models on RewardBench.
Furthermore, we demonstrate that the additional information provided by the distributional estimates can be utilized in downstream applications, such as risk-aware reinforcement learning, resulting in LLM policies that generate fewer extremely negative responses.
Our code and model are released at \url{https://github.com/Nicolinho/QRM}.
\end{abstract}

\section{Introduction}

Large Language Models (LLMs) have revolutionized natural language processing, demonstrating remarkable capabilities across a wide range of  tasks \citep{Anthropic@claude,team2023gemini,OpenAI2023GPT4TR}. However, the sheer scale and breadth of their training data present both opportunities and challenges. While LLMs can process and generate human-like text with unprecedented fluency, their outputs may not always align with human preferences, ethics, or real-world applicability.
To bridge this gap and ensure that these powerful tools truly benefit humanity, it has been recognized that fine-tuning techniques are necessary that can align LLMs with human values and intentions \citep{christiano2017deep,ziegler2019fine,bai2022training}. 
This process of refinement is essential to harness the full potential of LLMs while mitigating potential risks associated with their deployment in real-world scenarios.

Reinforcement learning from human feedback (RLHF) has emerged as a prominent and effective method to align LLMs with human preferences. 
RLHF uses reinforcement learning (RL) to fine-tune language models by maximizing rewards derived from a trained reward model.
This reward model, is itself learned from human preferences. 
By quantifying human judgments on responses for a prompt, the reward model provides a crucial bridge between human values and the optimization objective of the language model. 
As a result, accurate reward models are very important in order to finetune LLMs and it has been shown that improvements in reward model quality translates to improvements in the quality of the finetuned LLM \citep{touvron2023llama}.

However, current reward models are designed to output a single scalar value for a given query-response pair, an approach that fails to capture the inherent complexity and diversity of human values. This oversimplification can lead to problematic outcomes. For instance, in scenarios where human opinions diverge significantly - with some individuals finding a response appropriate while others deem it inappropriate - the reward model may resort to outputting an average value to minimize penalties across these disparate groups during training. 
This compromise fails to represent the nuanced spectrum of human preferences accurately. Moreover, it can hinder the learning process of the reward model if the training data contains conflicting labels where one kind of response is labeled as preferred in one scenario and as not-preferred in another similar scenario, potentially by different annotators.
It has been demonstrated that accurate preferences are crucial for developing robust reward models, with 'incorrect' preferences potentially confusing the model during training. However, in cases of conflicting preferences, the concept of 'correctness' becomes ambiguous, as these divergent views often represent equally valid but different perspectives rather than clear-cut right or wrong answers. This nuance poses a significant challenge to the current paradigm of reward modeling in RLHF.

\begin{figure*}[t]
\centering 
\includegraphics[width=0.7\textwidth]{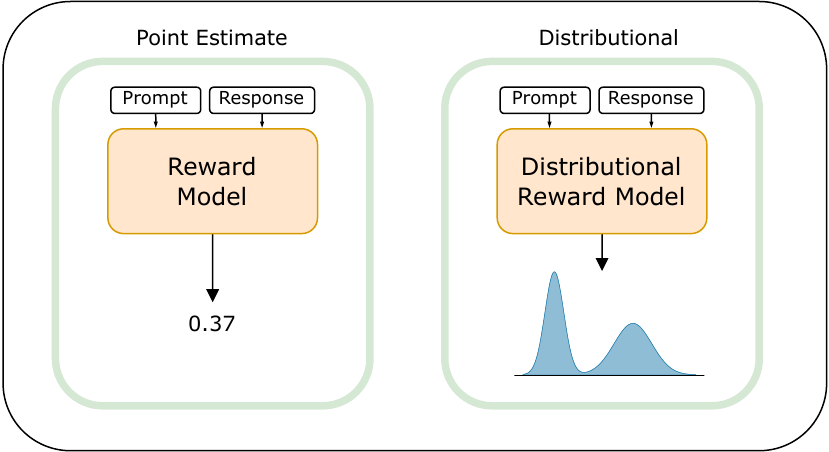}
\caption{Visualization of two reward estimation methods. The classical point estimate model generates a single scalar value as the reward. In contrast, the distributional reward model proposed in this paper outputs a distribution over possible rewards that can be multimodal.}
\label{fig:overview}
\end{figure*}

To address these limitations, we propose a novel approach to reward modeling that reimagines how human preferences are captured and represented. 
Instead of learning a single scalar reward for each query-response pair, our method, called Quantile Reward Model (QRM), learns a distribution over rewards. 
Specifically, we employ quantile regression to learn this distribution, allowing us to estimate different percentiles of the reward function and thus capture its entire shape.
This probabilistic approach offers a more nuanced and comprehensive representation of human values, capable of capturing the diversity and sometimes conflicting nature of preferences. 
By modeling a full distribution, we can account for the diverse values and preferences within a population without resorting to oversimplified averages.
Furthermore, it addresses the challenge of learning from seemingly conflicting preferences by treating them not as contradictions to be resolved, but as different modes in a multi-modal distribution. 
This shift in perspective allows the model to learn from a wider range of human feedback without the risk of confusion from what might previously have been considered 'incorrect' or conflicting inputs. 
Consequently, our distribution-based reward model promises to provide a more accurate, flexible, and robust foundation for reinforcement learning in language models, potentially leading to AI systems that are better aligned with the full spectrum of human values.

We implement our approach using a pretrained Llama-3 model \cite{meta_llama3} with 8 billion parameters. Our distributional reward model outperforms comparable methods on RewardBench \cite{lambert2024rewardbench}, a benchmark that evaluates reward models using a point estimate-based metric. 
This suggests that our model is better at handling conflicting values and the associated label noise during training.

Moreover, the additional information provided by the distributional estimate can be leveraged in downstream applications. 
We demonstrate this by training an RL policy with a risk-aware utility, derived from the distributional estimates of our model. 
The resulting policy achieves strong results and generates fewer extreme negative responses compared to a baseline trained using a risk-neutral point estimate as the reward.

\section{Related Work}

RLHF involves using RL to align language models with human preferences or feedback \citep{christiano2017deep,ziegler2019fine,stienon2020learning}. 
The process generally involves training a reward model on preference data collected from crowdworkers \citep{bai2022training, ouyang2022training} or model-selected responses \citep{bai2022constitutional}. Once a reward model is developed, RL algorithms can be used to train policies.
Another method is to directly optimize a policy by comparing selected and rejected responses using DPO \citep{rafailov2023direct} and followup works \citep{tang2024generalized}. 
While this method bypasses the need for an explicit reward model, our focus is on approaches that explicitly train such a model.

Estimating distributions rather than point estimates in regression has a long history and has proven valuable across many fields \citep{kneib2023rage}. Quantile regression \citep{Koenker_2005}, in particular, is a powerful tool for approximating distributions. In RL, distributional regression has also demonstrated its effectiveness \citep{bellemare2017distributional}, with quantile regression achieving especially strong results \citep{dabney2018distributional, tqc, dorka2022adaptively}.

A key distinction exists between epistemic and aleatoric uncertainty \citep{fox2011distinguishing,hullermeier2021aleatoric}. Epistemic uncertainty arises from incomplete knowledge and can be reduced by improving data or models. In contrast, aleatoric uncertainty stems from inherent randomness in the system and cannot be reduced. This paper focuses on modeling aleatoric uncertainty.

Several previous works have modeled preferences using distributions. Distributional Preference Learning (DPL) \citep{siththaranjan2023distributional} suggests that hidden contextual factors influence preferences, arising from the diverse values of annotators. This variability is captured by estimating a distribution over rewards. Similarly, the Distributional Preference Reward Model (DPRM) \citep{li2024aligning} learns a distribution over rewards, utilizing an optimal transport loss to train the model on preference data.
Both DPL and DPRM, like our approach, model a distribution over rewards. However, unlike our method, which incorporates attribute regression, these models rely solely on preference data and optimize the likelihood that the preferred response is ranked higher.

\section{Distributional Reward Models}

In this section, we first introduce the reinforcement learning from human feedback (RLHF) framework, then introduce multi-attribute regression reward models, followed by an explanation of quantile regression.
Finally, we propose our approach using quantile regression to obtain distributional reward models and explain how it can be used for risk-aware RLHF.

\subsection{Reinforcement Learning from Human Feedback}

The typical RLHF process using an explicit reward model consists of three stages:
\subsubsection{Supervised Fine-tuning}
In the first stage, the pre-trained language model is instruction-tuned by supervised learning on a dataset consisting of a prompt and a high-quality response. 
Similarly to the pre-training phase the model is trained with a cross-entropy loss over tokens but only on the tokens of the response. 
The resulting model is often used as the initialization for the reward model and the final policy trained with reinforcement learning.

\subsubsection{Reward Model Training}

After supervised fine-tuning, the next step is to train a reward model $r_\phi(x,y)$, which evaluates the quality of responses $y$ relative to the prompt $x$ based on human preferences.
Typically, reward models are trained on a preference dataset $D_{pref} = \{x, y^-, y^+\} $ consisting of a prompt $x$ and a preferred $y^+$ and not-preferred $y^-$ response.
The goal is that the reward model learns to assign higher rewards to outputs that better align with human preferences, such that it can later be used as an informative proxy signal in guiding the optimization of the language model.

The reward model is traditionally trained in accordance with the Bradley-Terry model.
The Bradley-Terry model posits that the probability of \(x, y^+ \) being preferred over \(x, y^- \) is given by:

\[
P(y^+ \succ y^-) = \frac{\exp(r(y^+))}{\exp(r(y^+)) + \exp(r(y^-))},
\]
where we dropped the dependence on the prompt $x$ for notational convenience.
The reward model can then be trained to minimize the negative log-likelihood of $y^+$ being preferred over $y^-$:

\[
\mathcal{L}_{\text{BT}}(\phi) = - \log P(y^+ \succ y^-),
\]

which, after substituting  can be written as:

\[
\mathcal{L}_{\text{BT}}(\phi) = - r_\phi(y^+) + \log \left(\exp(r_\phi(y^+)) + \exp(r_\phi(y^-))\right).
\]

\subsubsection{Reinforcement Learning}

Once the reward model is trained, it can be used to optimize the language model further through reinforcement learning. In this stage, the language model generates outputs for a dataset of prompts, and the reward model assigns rewards to these outputs based on their estimated quality. The language model is then trained to maximize these rewards.
This iterative process allows the model to improve its performance by continuously refining its behavior according to the feedback provided by the reward model, making it more capable of generating high-quality, human-aligned responses.

More formally, the goal is to finetune the language model \( \pi_\theta \) by maximizing the expected reward. To prevent reward hacking, which results in gibberish outputs, the model is penalized for deviating too much a reference policy. This is achieved with a KL divergence penalty between the current policy and the reference policy. Often the initial policy $\pi^{\text{sft}}$ is used as reference policy.  
The complete RL objective can be expressed as:

\[
\mathcal{L}(\theta) = \mathbb{E}_{x \sim \mathcal{D}, y \sim \pi_\theta(y|x)} \left[ r_\phi(x, y)  - \beta \, D_{\text{KL}}(\pi_\theta \| \pi_{\text{ref}}) \right],
\]

where \( \beta \) is a hyperparameter that controls the strength of the KL penalty, balancing the trade-off between maximizing the reward and staying close to the original model distribution.
In principle any RL algorithm can be used to optimize the policy. A common choice is to use PPO \citep{schulman2017proximal} and more recently more simple REINFORCE based algorithms \citep{ahmadian2024back}.

\subsection{Multi-Attribute Reward Models}
Most existing reward models are trained using the Bradley-Terry loss on binary preference data, where one response is labeled as preferred to another. This approach essentially frames the problem as binary classification. 
This method, however, fails to account for whether a response was clearly better or only marginally superior to the other. As a result, training a reward model on such data can lead to problems, as it may penalize good responses that are only slightly less favorable than an even better one in the same way it penalizes a very bad response.

Recent datasets are increasingly generated by first collecting absolute labels rather than relative ones \citep{cui2023ultrafeedback, wang2023helpsteer}. Responses are rated across various dimensions such as helpfulness, truthfulness, and harmlessness, with a fine-grained score assigned to each dimension. These individual scores are then aggregated to produce a final score. Preferences are subsequently determined by labeling the response with the higher aggregated score as the preferred one. 
Alongside these datasets, new approaches have emerged that use regression to directly estimate the fine-grained scores \citep{wang2023helpsteer,wang2024armo}. 
In this framework, the reward network outputs a single scalar value for each objective and is trained to minimize the mean squared error between its predictions and the fine-grained scores.

For prompts $x$, responses $y$, $M \in \mathbb{N}$ attributes with corresponding scores $l_i$, $i \in [1,...,M]$  the optimization objective becomes
\begin{equation}
\min_{\theta} \sum_{x,y,l_i \in \mathcal{D}} \sum_{i=1}^M || f_\theta(x,y)_i - l_i ||^2_2,
\end{equation}
where $f_\theta$ is the reward model with $M$ outputs and with parameters $\theta$.

Finally, a single reward score is derived by aggregating the individual scores using a weighted sum. The weights can either be predefined \citep{wang2024helpsteer2} or learned through a gating network. 
For instance, ArmoRM \citep{wang2024armo} optimizes the gating network using the Bradley-Terry loss on preference data, where the final rewards are aggregated from individual attribute scores based on the gating weights. 
During this optimization, the regression network remains frozen. 
Methods using this approach of first estimating attribute scores have demonstrated significant success, achieving top results on the RewardBench leaderboard \citep{lambert2024rewardbench}.

\subsection{Quantile Regression}

Quantile regression \citep{Koenker_2005} is a versatile statistical technique that generalizes the conventional linear regression model by estimating the conditional quantiles of the response variable. Formally, while ordinary least squares (OLS) regression focuses on minimizing the sum of squared residuals to estimate the conditional mean, quantile regression minimizes an asymmetric loss function to estimate specific quantiles belonging to the distribution of the response variable.
The quantile function $Q(\tau)$ (where $0<\tau<1$) of a probability distribution is the inverse of its cumulative distribution function (CDF) and hence  $Q(\tau)$ denotes  the value $x$ such that the probability of the corresponding random variable $X$ being lower than $x$ is equal to $\tau$, i.e. $P(X <= x) = \tau$.
For a given quantile $\tau$ ,  linear quantile regression solves the following optimization problem:

\begin{equation}
\min_{w} \sum_{i: y_i \geq x_i^\top w} \tau |y_i - x_i^\top w| + \sum_{i: y_i < x_i^\top w} (1-\tau) |y_i - x_i^\top w|.
\end{equation}

Here, $y_i \in \mathbb{R}$ represents the labels, $x_i  \in \mathbb{R}^d$ the d-dimensional inputs, and $w  \in \mathbb{R}^d$ is the vector of coefficients to be estimated.

One of the key strengths of quantile regression lies in its ability to model the conditional distribution of the response variable at different quantiles, offering a more detailed view of the relationship between the dependent and independent variables.
This is particularly advantageous in situations where the effects of the input variables are not uniform across the distribution of the outcome variable, such as in the presence of skewed or multimodal distributions, heteroscedasticity, or outliers. 
Unlike OLS, which provides a single estimate of the central tendency, quantile regression allows to approximately represent a distribution over the response variable, offering more information about the data.
Hence, quantile regression provides a comprehensive framework for understanding the distributional effects of input variables on the response variable, making it a critical tool in applications where understanding the full conditional distribution is essential.

\subsection{Disrtibutional Reward Models via Quantile Regression}

\begin{figure*}[t]
\centering 
\includegraphics[width=\textwidth]{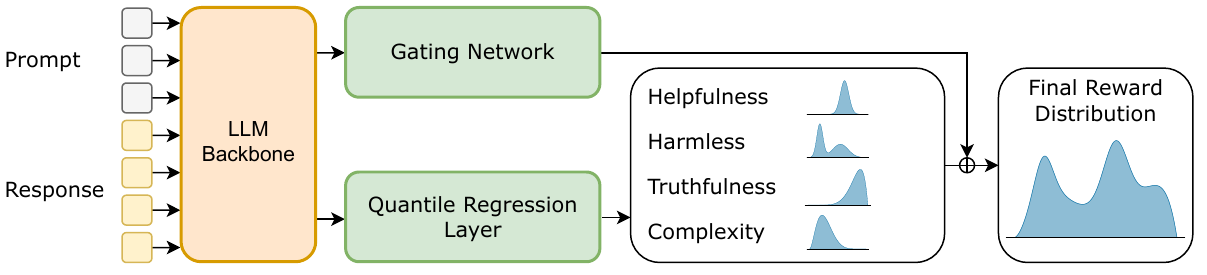}
\caption{
Visualization of our approach. The prompt and response are fed into the LLM backbone, which generates two types of embeddings: a prompt embedding for the gating network and a prompt-response embedding for the quantile regression layers. The quantile regression layers produce quantile estimates for various attributes, such as helpfulness and harmlessness. Simultaneously, the gating network computes weights for the individual distributions corresponding to these quantile estimates. The final output is a mixture distribution, formed by combining the attribute distributions weighted by the outputs of the gating network.}
\label{fig:approach}
\end{figure*}

Our objective is to develop a reward model that outputs a distribution over rewards. To achieve this, we propose a two-step approach called Quantile Reward Models (QRM). This method involves: (1) estimating distributions over attribute scores (e.g., helpfulness and harmlessness) using quantile regression, and (2) training a gating network to aggregate these individual attribute distributions into a final reward distribution.
An illustration of our approach is depicted in Figure \ref{fig:approach}.

\subsubsection{Step 1: Attribute Distribution Estimation}
In the first step, we estimate the distributions for each of the $M$ attributes using quantile regression. Specifically, we perform regression on $K \in \mathbb{N}$ evenly spaced quantiles $\tau_k \in (0,1)$. For each attribute, we train $K$ linear quantile regression layers, with each layer predicting the value at a specific quantile.

The input to each quantile regression layer is a feature vector derived from a frozen LLM backbone. These feature vectors remain fixed throughout training. This phase requires datasets that provide absolute scores for each attribute to allow for distribution estimation.

\subsubsection{Step 2: Gating Network for Distribution Aggregation}
In the second step, we train a gating network that predicts the mixing weights for combining the individual attribute distributions into a final reward distribution. The gating network outputs weights $g_m$ for each attribute $m=1,...,M$, ensuring that they sum to one: $\sum_m g_m = 1$.
These weights determine the contribution of each attribute's distribution to the final distribution. The mixed distribution is then computed as the weighted sum of the $M$ quantile values for each attribute, given by:
\begin{equation}
    Q_{\text{mix}}(\tau_i)  = \sum_{m=1}^M g_m \cdot Q_m (\tau_i).
\end{equation}
This calculation is performed for all quantiles $\tau_i$.
The input to the gating network is a feature vector from the LLM backbone that encodes only the prompt.
The gating network is trained using preference data and the Bradley-Terry loss following the approach in \citep{wang2024armo}. To adapt this approach to our scenario, we compute the expected value of the final distribution, which serves as the reward for optimizing the Bradley-Terry loss. Importantly, only the gating network is trained during this stage, while the quantile regression layers remain fixed.

In the end, we obtain both a distribution over attributes and a final distribution over rewards. Notably, our approach can also accommodate fixed attribute weightings, as has been done in works such as HelpSteer \citep{wang2023helpsteer, wang2024helpsteer2}.

\subsection{Risk-Aware Reinforcement Learning from Human Feedback}

We believe that distributional reward models offer significant potential by leveraging the additional information contained in the reward distribution for downstream reinforcement learning (RL) tasks. In this section, we demonstrate one possible application of this concept.

When deploying a chat model, a critical concern is avoiding particularly poor responses, as low-quality outputs can negatively impact user experience across various dimensions, such as overall quality and safety. Even occasional bad responses can lead to user dissatisfaction. In these cases, it can be beneficial to optimize a metric that emphasizes penalizing low-quality responses as perceived by users.

A point estimate reward model provides only a single value to assess a response. For example, if a response is perceived as excellent by some users and poor by others, a point estimate model may produce an average score that can lead the LLM to generate such mixed-quality responses. In contrast, a distributional reward model allows us to penalize responses with significant disagreement.

One way to achieve this is by applying a concave utility function over the reward distribution to bias the final score to emphasize lower values, i.e., the left tail of the distribution. We define this utility function as:
\begin{equation}
    \text{Utility} = \mathbb{E}_{r \sim \mathcal{P}} \left[ - e^{- \lambda r} \right],
\end{equation}
where $\lambda$ is a hyperparameter controlling the emphasis on low rewards, and the expectation is taken over the distribution $\mathcal{P}$ of reward values $r$ produced by our distributional reward model. We can approximate this by applying the utility function to each quantile estimate before computing the final expectation.

This utility function penalizes distributions with substantial probability mass on low reward values. For instance, a bimodal distribution with one peak at a low value and another at a high value will have a lower utility than an unimodal distribution centered around the same expected value. Consequently, this approach encourages risk-averse policies that avoid producing highly variable or risky responses.

\section{Experiments}

\subsection{Implementation}

Our LLM backbone is based on LLaMA-3 with 8 billion parameters \citep{meta_llama3}, initialized with weights from a reward model trained using the Bradley-Terry loss \citep{dong2024rlhf}. The backbone remains frozen during both stages of training. This allows for significant computational efficiency, as we can precompute the backbone features once and reuse them throughout the subsequent stages, thereby significantly reducing the required computation resources.
For multi-attribute regression, we follow prior work and use $19$ attributes from \cite{wang2024armo}, sourced from eight datasets (details provided in the appendix). To limit the computational requirements, we limit the number of data points per attribute to $60,000$ when training the linear quantile regression layers.
The quantile regression is implemented using Scikit-learn \cite{scikit-learn}, with L1 weight regularization set to $0.003$. For each attribute, we train $19$ quantile regression models corresponding to the evenly spaced quantiles: $0.05, 0.10, \dots, 0.90, 0.95$. Additionally, as in \cite{wang2024armo}, we mitigate the correlation between the verbosity attribute and other attributes by applying their penalty scores, and adjusting the quantile estimates accordingly.

The gating network used for aggregation is a multi-layer perceptron with three hidden layers and a softmax activation at the output layer. We train the network for $3$ epochs using the AdamW \citep{adamw} optimizer with a learning rate of $0.0003$, batch size of $1024$, and a cosine decay learning rate scheduler. The weight decay is set to $0.001$. The training is performed on data from $10$ datasets, with further details outlined in the appendix.

\subsection{ RewardBench Results}
\begin{table*}[t]
		\begin{tabular}{ll|c|cccc}
			\toprule
			\multicolumn{1}{c}{\textbf{Method}} &
			\multicolumn{1}{c}{\textbf{Base Model}} &
			\multicolumn{1}{c}{\textbf{Score}} &
			\multicolumn{1}{c}{\textbf{Chat}} &
			\multicolumn{1}{c}{\textbf{Chat Hard}} &
			\multicolumn{1}{c}{\textbf{Safety}} &
			\multicolumn{1}{c}{\textbf{Reasoning}} \\
			\midrule
			HelpSteer2 RM & Nemotron-4 340B & \textbf{92.2} &  95.8  & \textbf{87.1} & 91.5 & 93.7\\
			\midrule
			ArmoRM   & Llama-3 8B  & 90.8   & 96.9 & 76.8 & 92.2 & 97.3 \\
			HelpSteer2 RM & Llama-3 70B & 86.3 &  91.3 & 80.3 & \textbf{92.8} & 90.7  \\
			LLM-as-a-judge   & GPT-4 Turbo & 84.2 & 95.3 & 74.3 & 87.2 & 86.9  \\
			LLM-as-a-judge   & GPT-4o  & 83.3 & 96.6 & 70.4 & 86.7 & 84.9 \\
			Bradley-Terry    & Llama-3 8B  & 83.6 & \textbf{99.4} & 65.1 & 87.8 & 86.4  \\
			\midrule
			\textbf{QRM}     & Llama-3 8B  & \textbf{91.2}   & 97.2 & 78.5 & 91.1 & \textbf{98.0} \\
			\bottomrule
		\end{tabular}
	\caption{Performance comparison on RewardBench. The benchmark consists of four categories. The overall score is computed as the weighted average over the single scores. A higher value is better.}
	\label{tab:reward_bench}
	\vspace{-0.8em}
\end{table*}
To examine the general capabilities of our distributional reward model we evaluate it on RewardBench \cite{lambert2024rewardbench} which is a benchmark for evaluating reward models. It features diverse tasks across 4 categories: Chat, Chat Hard, Safety, and Reasoning. Each category includes datasets with pairwise preference data (chosen vs. rejected responses), and the final score is a weighted average across categories.
Following prior work \citep{kim2024prometheus, wang2024helpsteer2} we do not evaluate the fifth category of the benchmark named Prior Sets as there are several flaws with this subset \citep[p.~25]{wang2024helpsteer2}.

We compare our method to various prior methods:  HelpSteer2 model with both the Nemotron-4 340B and the LLama-3 70B base model \cite{wang2024helpsteer2},  ArmoRM  \cite{wang2024armo}, FsfairX-LLaMA3-RM-v0.1 \cite{dong2024rlhf} which was trained with the Bradley-Terry loss and is used as the backbone for our approach, using GPT4 as a judge \cite{zheng2023judging}.
The comparison to ArmoRM is the most interesting as ArmoRM uses a similar approach but with a point estimate for every attribute.
We show the results of the evaluation in Table \ref{tab:reward_bench}.
The results show that QRM achieves the highest performance among models with the same base model.
Most notably, it achieves a slightly higher performance than ArmoRM. 
We suspect a reason is that our model can better handle conflicting annotations during training without confusing the model.
Only Nemotron-4 with a much larger model of $340$ billion parameters achieves a higher performance than our model.
However, we note that the main advantage of our approach does not come when evaluated with a metric based on final point estimate but when the additional information is used in a downstream process.

\subsection{Risk-Aware RLHF Experiments}

\begin{figure}[t]
    \centering
    \begin{subfigure}[b]{0.325\textwidth}
        \centering
        \includegraphics[width=\textwidth]{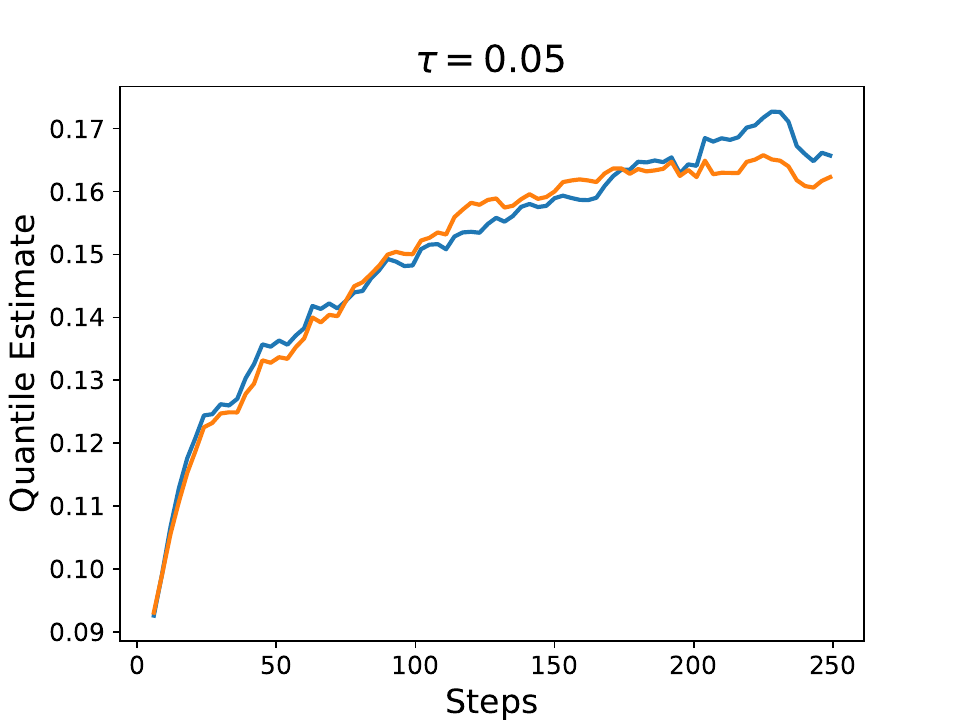}
    \end{subfigure}
    \begin{subfigure}[b]{0.325\textwidth}
        \centering
        \includegraphics[width=\textwidth]{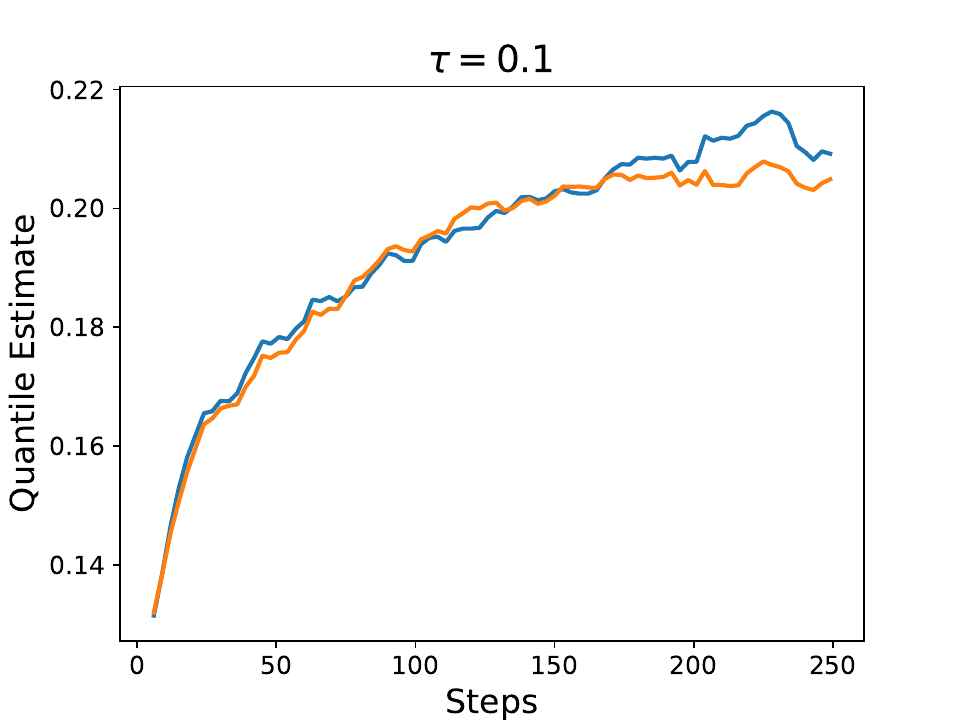}
    \end{subfigure}
    \begin{subfigure}[b]{0.325\textwidth}
        \centering
        \includegraphics[width=\textwidth]{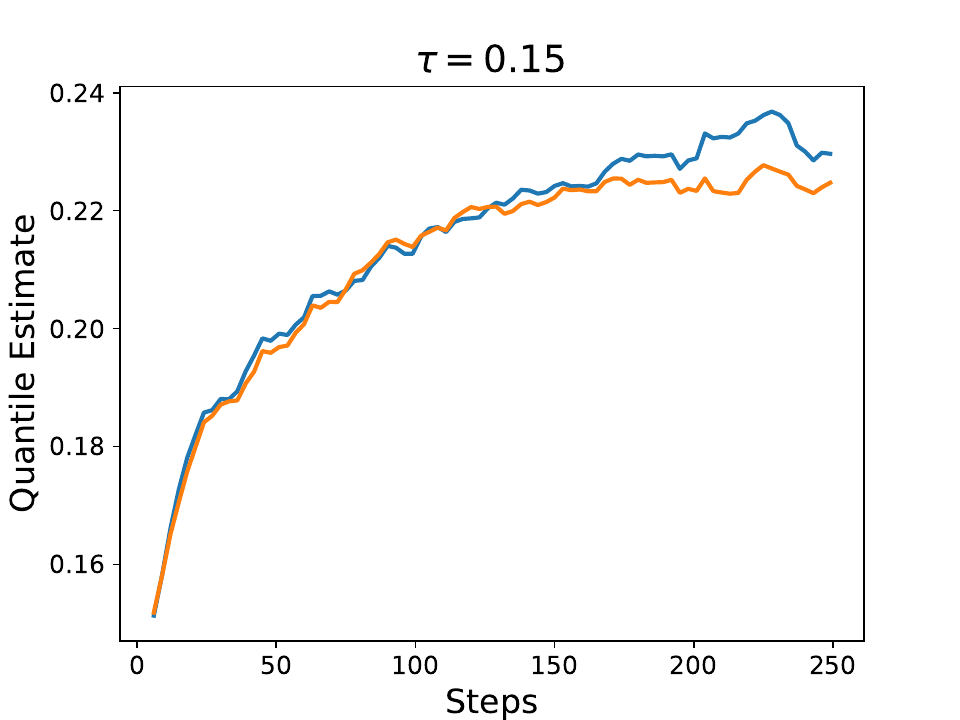}
    \end{subfigure}

    \begin{subfigure}[b]{0.325\textwidth}
        \centering
        \includegraphics[width=\textwidth]{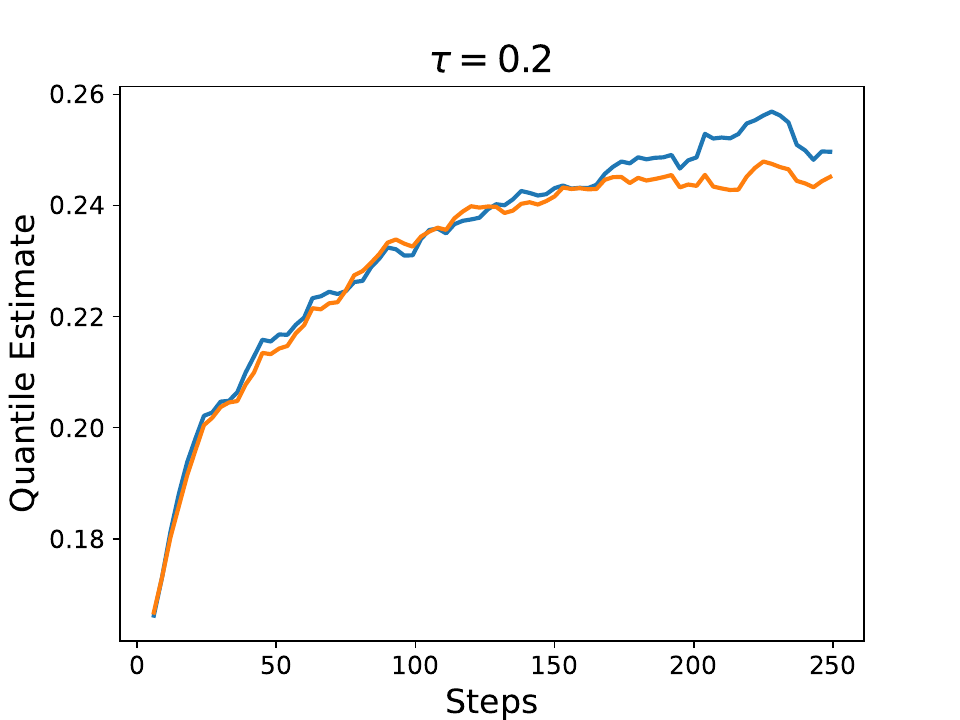}
    \end{subfigure}
    \begin{subfigure}[b]{0.325\textwidth}
        \centering
        \includegraphics[width=\textwidth]{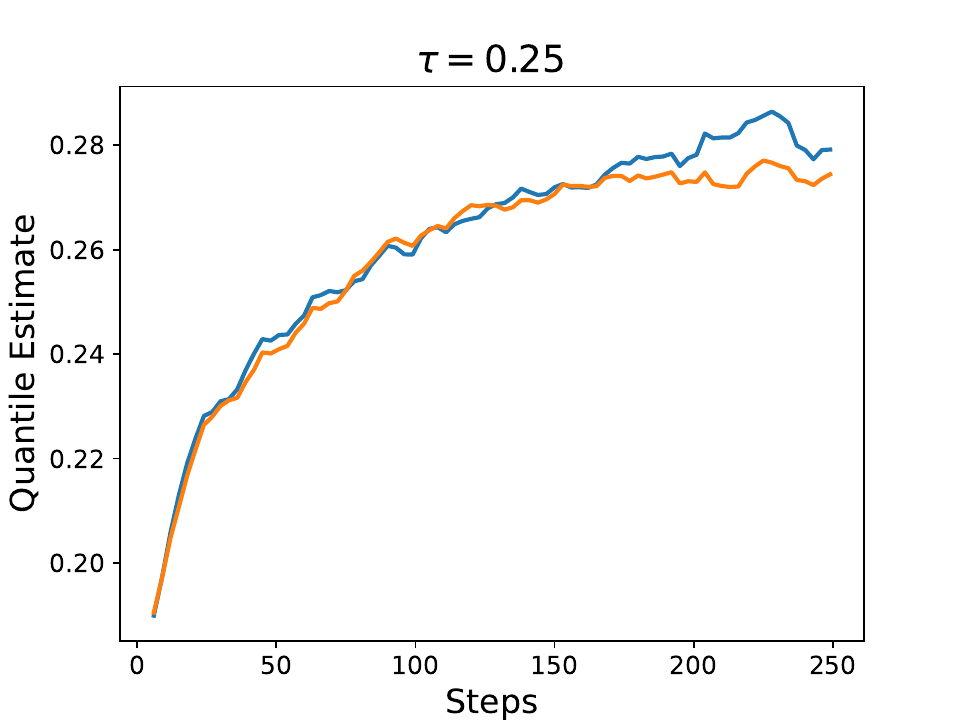}
    \end{subfigure}
    \begin{subfigure}[b]{0.325\textwidth}
        \centering
        \includegraphics[width=\textwidth]{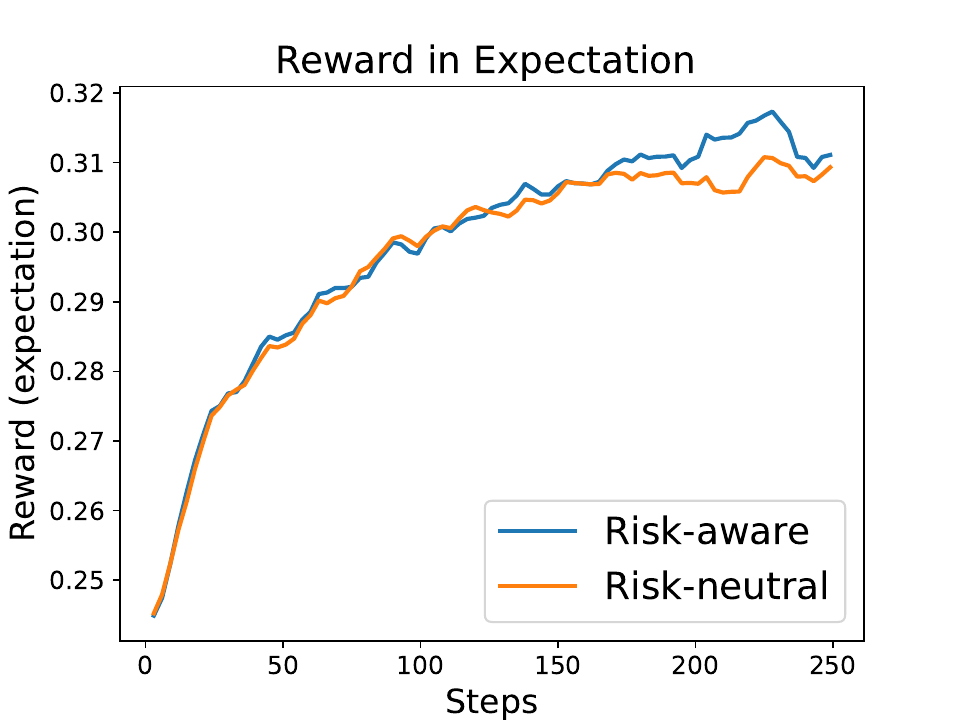}
    \end{subfigure}
    
    \caption{Results for two policies trained with RL using QRM as reward model. One policy was trained with the expectation as reward (risk-neutral) while the other was trained with a risk-aware utility function. We show the quantile estimates for the left tail quantiles $\tau =0.05,...,0.25$ and the expectation of the reward.} 

\label{fig:rlhf_experiment}
\end{figure}

To evaluate our approach, we train two RL policies. For comparability, both use our distributional reward model, but in one case we apply the risk-aware utility function to compute the rewards, while in the other, we simulate a point-estimate reward by using the expected value of the reward distribution.

We use the LLaMA-3 model with 8 billion parameters as the base model, training on the Anthropic HH-RLHF dataset \citep{bai2022training}. Both supervised fine-tuning (SFT) and RL are performed using LoRA, with a rank value of $32$ and $\alpha=64$. For SFT, we train on the responses from the dataset selected as chosen, filtering for examples with at most $1024$ tokens. The SFT model is trained for one epoch with a learning rate of $0.0003$, batch size of $1024$, using the AdamW optimizer and a cosine learning rate schedule.

After completing the SFT stage, we merge the LoRA weights into the model's base weights. For RL, we optimize the LLM policy using the RLOO algorithm \citep{ahmadian2024back}, initializing both the policy and reference policy with the SFT model. We train with a value of $k=2$ for RLOO, a fixed learning rate of $0.0001$, and a batch size of $1024$, using AdamW. We filter prompts to those with fewer than $348$ tokens and limit the response length to $512$ tokens.

We present the results of our experiment in Figure \ref{fig:rlhf_experiment}. The figure illustrates the development of quantile estimates from our distributional reward model for quantiles $\tau = 0.05, \ldots, 0.25$, which represent the left tail of the distribution.
The results indicate that the risk-aware utility leads to higher final values for these quantiles compared to the risk-neutral utility. This suggests that the final policy is more risk-aware, resulting in fewer responses that are likely to be deemed very poor or inappropriate.
Additionally, we examine the performance of both policies in terms of the expected reward, a risk-neutral metric. Both policies achieve similar final performance on this metric, demonstrating that employing the risk-aware utility does not lead to an overall decrease in performance.

\section{Conclusion}

In this work, we proposed QRM, a novel approach for training a distributional reward model in the context of RLHF. Traditionally, reward models are treated as point estimators, which face challenges in accommodating diverse human preferences and handling the resulting label noise during training. In contrast, estimating a distribution that can be multimodal is more powerful, capturing distinct values and preferences more effectively.
QRM uses quantile regression to estimate distributions over attribute scores such as helpfulness and harmlessness. A gating network produces weights, aggregating these individual distributions into a final reward distribution.

Our experimental results demonstrate that QRM outperforms previous approaches with comparable base architectures. 
Notably, even when reduced to point estimates by calculating the expectation of the distribution, our model showed improvements.
Moreover, the distributional reward model provides additional information that can be utilized for risk-aware RLHF. 
The final policy generated by our approach produces responses with higher quantile estimates for the left tail of the distribution, reducing the occurrence of extremely poor responses.

Our work opens several promising avenues for future research. 
One exciting direction is to directly incorporate conflicting labels into training. Unlike traditional preference learning datasets that rely on majority-vote labels, our model could account for raw, conflicting annotations for the same prompt-response pair. Additionally, it would be interesting to explore integrating our distributional reward model with distributional RL algorithms that estimate a distribution for the value function. 
Another promising avenue is to leverage the extra information provided by our model within search algorithms during decoding.
More broadly, the extra information provided by our distributional reward model allows for more interpretability as well as more creative ways to steer LLMs during downstream fine-tuning.

\newpage







\bibliography{iclr2025_conference}

\begin{thebibliography}{42}
\providecommand{\natexlab}[1]{#1}
\providecommand{\url}[1]{\texttt{#1}}
\expandafter\ifx\csname urlstyle\endcsname\relax
  \providecommand{\doi}[1]{doi: #1}\else
  \providecommand{\doi}{doi: \begingroup \urlstyle{rm}\Url}\fi

\bibitem[Ahmadian et~al.(2024)Ahmadian, Cremer, Gall{\'e}, Fadaee, Kreutzer,
  {\"U}st{\"u}n, and Hooker]{ahmadian2024back}
Arash Ahmadian, Chris Cremer, Matthias Gall{\'e}, Marzieh Fadaee, Julia
  Kreutzer, Ahmet {\"U}st{\"u}n, and Sara Hooker.
\newblock Back to basics: Revisiting reinforce style optimization for learning
  from human feedback in llms.
\newblock \emph{arXiv preprint arXiv:2402.14740}, 2024.

\bibitem[Anthropic(2023)]{Anthropic@claude}
Anthropic.
\newblock Introducing claude.
\newblock 2023.
\newblock URL \url{https://www.anthropic.com/index/introducing-claude}.

\bibitem[Bai et~al.(2022{\natexlab{a}})Bai, Jones, Ndousse, Askell, Chen,
  DasSarma, Drain, Fort, Ganguli, Henighan, et~al.]{bai2022training}
Yuntao Bai, Andy Jones, Kamal Ndousse, Amanda Askell, Anna Chen, Nova DasSarma,
  Dawn Drain, Stanislav Fort, Deep Ganguli, Tom Henighan, et~al.
\newblock Training a helpful and harmless assistant with reinforcement learning
  from human feedback.
\newblock \emph{arXiv preprint arXiv:2204.05862}, 2022{\natexlab{a}}.

\bibitem[Bai et~al.(2022{\natexlab{b}})Bai, Kadavath, Kundu, Askell, Kernion,
  Jones, Chen, Goldie, Mirhoseini, McKinnon, et~al.]{bai2022constitutional}
Yuntao Bai, Saurav Kadavath, Sandipan Kundu, Amanda Askell, Jackson Kernion,
  Andy Jones, Anna Chen, Anna Goldie, Azalia Mirhoseini, Cameron McKinnon,
  et~al.
\newblock Constitutional ai: Harmlessness from ai feedback.
\newblock \emph{arXiv preprint arXiv:2212.08073}, 2022{\natexlab{b}}.

\bibitem[Bellemare et~al.(2017)Bellemare, Dabney, and
  Munos]{bellemare2017distributional}
Marc~G Bellemare, Will Dabney, and R{\'e}mi Munos.
\newblock A distributional perspective on reinforcement learning.
\newblock In \emph{International Conference on Machine Learning}, pp.\
  449--458, 2017.

\bibitem[Christiano et~al.(2017)Christiano, Leike, Brown, Martic, Legg, and
  Amodei]{christiano2017deep}
Paul~F Christiano, Jan Leike, Tom Brown, Miljan Martic, Shane Legg, and Dario
  Amodei.
\newblock Deep reinforcement learning from human preferences.
\newblock \emph{Advances in neural information processing systems}, 30, 2017.

\bibitem[Cui et~al.(2023)Cui, Yuan, Ding, Yao, Zhu, Ni, Xie, Liu, and
  Sun]{cui2023ultrafeedback}
Ganqu Cui, Lifan Yuan, Ning Ding, Guanming Yao, Wei Zhu, Yuan Ni, Guotong Xie,
  Zhiyuan Liu, and Maosong Sun.
\newblock Ultrafeedback: Boosting language models with high-quality feedback,
  2023.

\bibitem[Dabney et~al.(2018)Dabney, Rowland, Bellemare, and
  Munos]{dabney2018distributional}
Will Dabney, Mark Rowland, Marc Bellemare, and R{\'e}mi Munos.
\newblock Distributional reinforcement learning with quantile regression.
\newblock In \emph{Proceedings of the AAAI Conference on Artificial
  Intelligence}, volume~32, 2018.

\bibitem[Daniele \& Suphavadeeprasit(2023)Daniele and
  Suphavadeeprasit]{daniele2023amplify-instruct}
Luigi Daniele and Suphavadeeprasit.
\newblock Amplify-instruct: Synthetically generated diverse multi-turn
  conversations for efficient llm training.
\newblock \emph{arXiv preprint arXiv:(coming soon)}, 2023.
\newblock URL \url{https://huggingface.co/datasets/LDJnr/Capybara}.

\bibitem[Dong et~al.(2024)Dong, Xiong, Pang, Wang, Zhao, Zhou, Jiang, Sahoo,
  Xiong, and Zhang]{dong2024rlhf}
Hanze Dong, Wei Xiong, Bo~Pang, Haoxiang Wang, Han Zhao, Yingbo Zhou, Nan
  Jiang, Doyen Sahoo, Caiming Xiong, and Tong Zhang.
\newblock Rlhf workflow: From reward modeling to online rlhf.
\newblock \emph{arXiv preprint arXiv:2405.07863}, 2024.

\bibitem[Dorka et~al.(2022)Dorka, Welschehold, B{\"o}decker, and
  Burgard]{dorka2022adaptively}
Nicolai Dorka, Tim Welschehold, Joschka B{\"o}decker, and Wolfram Burgard.
\newblock Adaptively calibrated critic estimates for deep reinforcement
  learning.
\newblock \emph{IEEE Robotics and Automation Letters}, 8\penalty0 (2):\penalty0
  624--631, 2022.

\bibitem[Ethayarajh et~al.(2022)Ethayarajh, Choi, and Swayamdipta]{SHP}
Kawin Ethayarajh, Yejin Choi, and Swabha Swayamdipta.
\newblock Understanding dataset difficulty with $\mathcal{V}$-usable
  information.
\newblock In Kamalika Chaudhuri, Stefanie Jegelka, Le~Song, Csaba Szepesvari,
  Gang Niu, and Sivan Sabato (eds.), \emph{Proceedings of the 39th
  International Conference on Machine Learning}, volume 162 of
  \emph{Proceedings of Machine Learning Research}, pp.\  5988--6008. PMLR,
  17--23 Jul 2022.

\bibitem[Fox \& {\"U}lk{\"u}men(2011)Fox and
  {\"U}lk{\"u}men]{fox2011distinguishing}
Craig~R Fox and G{\"u}lden {\"U}lk{\"u}men.
\newblock Distinguishing two dimensions of uncertainty.
\newblock \emph{Fox, Craig R. and G{\"u}lden {\"U}lk{\"u}men
  (2011),“Distinguishing Two Dimensions of Uncertainty,” in Essays in
  Judgment and Decision Making, Brun, W., Kirkeb{\o}en, G. and Montgomery, H.,
  eds. Oslo: Universitetsforlaget}, 2011.

\bibitem[Ganguli et~al.(2022)Ganguli, Lovitt, Kernion, Askell, Bai, Kadavath,
  Mann, Perez, Schiefer, Ndousse, Jones, Bowman, Chen, Conerly, DasSarma,
  Drain, Elhage, El-Showk, Fort, Hatfield-Dodds, Henighan, Hernandez, Hume,
  Jacobson, Johnston, Kravec, Olsson, Ringer, Tran-Johnson, Amodei, Brown,
  Joseph, McCandlish, Olah, Kaplan, and Clark]{ganguli2022red}
Deep Ganguli, Liane Lovitt, Jackson Kernion, Amanda Askell, Yuntao Bai, Saurav
  Kadavath, Ben Mann, Ethan Perez, Nicholas Schiefer, Kamal Ndousse, Andy
  Jones, Sam Bowman, Anna Chen, Tom Conerly, Nova DasSarma, Dawn Drain, Nelson
  Elhage, Sheer El-Showk, Stanislav Fort, Zac Hatfield-Dodds, Tom Henighan,
  Danny Hernandez, Tristan Hume, Josh Jacobson, Scott Johnston, Shauna Kravec,
  Catherine Olsson, Sam Ringer, Eli Tran-Johnson, Dario Amodei, Tom Brown,
  Nicholas Joseph, Sam McCandlish, Chris Olah, Jared Kaplan, and Jack Clark.
\newblock Red teaming language models to reduce harms: Methods, scaling
  behaviors, and lessons learned, 2022.

\bibitem[H{\"u}llermeier \& Waegeman(2021)H{\"u}llermeier and
  Waegeman]{hullermeier2021aleatoric}
Eyke H{\"u}llermeier and Willem Waegeman.
\newblock Aleatoric and epistemic uncertainty in machine learning: An
  introduction to concepts and methods.
\newblock \emph{Machine learning}, 110\penalty0 (3):\penalty0 457--506, 2021.

\bibitem[Ji et~al.(2023)Ji, Liu, Dai, Pan, Zhang, Bian, Chen, Sun, Wang, and
  Yang]{ji2023beavertails}
Jiaming Ji, Mickel Liu, Juntao Dai, Xuehai Pan, Chi Zhang, Ce~Bian, Boyuan
  Chen, Ruiyang Sun, Yizhou Wang, and Yaodong Yang.
\newblock Beavertails: Towards improved safety alignment of {LLM} via a
  human-preference dataset.
\newblock In \emph{Thirty-seventh Conference on Neural Information Processing
  Systems Datasets and Benchmarks Track}, 2023.
\newblock URL \url{https://openreview.net/forum?id=g0QovXbFw3}.

\bibitem[Kim et~al.(2024{\natexlab{a}})Kim, Shin, Cho, Jang, Longpre, Lee, Yun,
  Shin, Kim, Thorne, and Seo]{prometheus}
Seungone Kim, Jamin Shin, Yejin Cho, Joel Jang, Shayne Longpre, Hwaran Lee,
  Sangdoo Yun, Seongjin Shin, Sungdong Kim, James Thorne, and Minjoon Seo.
\newblock Prometheus: Inducing fine-grained evaluation capability in language
  models.
\newblock In \emph{The Twelfth International Conference on Learning
  Representations}, 2024{\natexlab{a}}.
\newblock URL \url{https://openreview.net/forum?id=8euJaTveKw}.

\bibitem[Kim et~al.(2024{\natexlab{b}})Kim, Suk, Longpre, Lin, Shin, Welleck,
  Neubig, Lee, Lee, and Seo]{kim2024prometheus}
Seungone Kim, Juyoung Suk, Shayne Longpre, Bill~Yuchen Lin, Jamin Shin, Sean
  Welleck, Graham Neubig, Moontae Lee, Kyungjae Lee, and Minjoon Seo.
\newblock Prometheus 2: An open source language model specialized in evaluating
  other language models, 2024{\natexlab{b}}.

\bibitem[Kneib et~al.(2023)Kneib, Silbersdorff, and S{\"a}fken]{kneib2023rage}
Thomas Kneib, Alexander Silbersdorff, and Benjamin S{\"a}fken.
\newblock Rage against the mean--a review of distributional regression
  approaches.
\newblock \emph{Econometrics and Statistics}, 26:\penalty0 99--123, 2023.

\bibitem[Koenker(2005)]{Koenker_2005}
Roger Koenker.
\newblock \emph{Quantile Regression}.
\newblock Econometric Society Monographs. Cambridge University Press, 2005.

\bibitem[Kuznetsov et~al.(2020)Kuznetsov, Shvechikov, Grishin, and Vetrov]{tqc}
Arsenii Kuznetsov, Pavel Shvechikov, Alexander Grishin, and Dmitry Vetrov.
\newblock Controlling overestimation bias with truncated mixture of continuous
  distributional quantile critics.
\newblock In \emph{International Conference on Machine Learning}, pp.\
  5556--5566. PMLR, 2020.

\bibitem[Lambert et~al.(2024)Lambert, Pyatkin, Morrison, Miranda, Lin, Chandu,
  Dziri, Kumar, Zick, Choi, et~al.]{lambert2024rewardbench}
Nathan Lambert, Valentina Pyatkin, Jacob Morrison, LJ~Miranda, Bill~Yuchen Lin,
  Khyathi Chandu, Nouha Dziri, Sachin Kumar, Tom Zick, Yejin Choi, et~al.
\newblock Rewardbench: Evaluating reward models for language modeling.
\newblock \emph{arXiv preprint arXiv:2403.13787}, 2024.

\bibitem[Li et~al.(2024)Li, Zhang, Dong, Deik, Tang, and Liu]{li2024aligning}
Dexun Li, Cong Zhang, Kuicai Dong, Derrick Goh~Xin Deik, Ruiming Tang, and Yong
  Liu.
\newblock Aligning crowd feedback via distributional preference reward
  modeling.
\newblock \emph{arXiv preprint arXiv:2402.09764}, 2024.

\bibitem[Lightman et~al.(2023)Lightman, Kosaraju, Burda, Edwards, Baker, Lee,
  Leike, Schulman, Sutskever, and Cobbe]{lightman2023let}
Hunter Lightman, Vineet Kosaraju, Yura Burda, Harri Edwards, Bowen Baker, Teddy
  Lee, Jan Leike, John Schulman, Ilya Sutskever, and Karl Cobbe.
\newblock Let's verify step by step.
\newblock \emph{arXiv preprint arXiv:2305.20050}, 2023.

\bibitem[Loshchilov \& Hutter(2019)Loshchilov and Hutter]{adamw}
Ilya Loshchilov and Frank Hutter.
\newblock Decoupled weight decay regularization.
\newblock In \emph{International Conference on Learning Representations}, 2019.
\newblock URL \url{https://openreview.net/forum?id=Bkg6RiCqY7}.

\bibitem[Meta(2024)]{meta_llama3}
Meta.
\newblock Introducing meta llama 3: The most capable openly available llm to
  date.
\newblock \emph{Meta AI Blog}, 2024.
\newblock \url{https://ai.meta.com/blog/meta-llama-3/}.

\bibitem[OpenAI(2023)]{OpenAI2023GPT4TR}
OpenAI.
\newblock Gpt-4 technical report.
\newblock \emph{ArXiv}, abs/2303.08774, 2023.

\bibitem[Ouyang et~al.(2022)Ouyang, Wu, Jiang, Almeida, Wainwright, Mishkin,
  Zhang, Agarwal, Slama, Ray, et~al.]{ouyang2022training}
Long Ouyang, Jeffrey Wu, Xu~Jiang, Diogo Almeida, Carroll Wainwright, Pamela
  Mishkin, Chong Zhang, Sandhini Agarwal, Katarina Slama, Alex Ray, et~al.
\newblock Training language models to follow instructions with human feedback.
\newblock \emph{Advances in Neural Information Processing Systems},
  35:\penalty0 27730--27744, 2022.

\bibitem[Pedregosa et~al.(2011)Pedregosa, Varoquaux, Gramfort, Michel, Thirion,
  Grisel, Blondel, Prettenhofer, Weiss, Dubourg, Vanderplas, Passos,
  Cournapeau, Brucher, Perrot, and Duchesnay]{scikit-learn}
F.~Pedregosa, G.~Varoquaux, A.~Gramfort, V.~Michel, B.~Thirion, O.~Grisel,
  M.~Blondel, P.~Prettenhofer, R.~Weiss, V.~Dubourg, J.~Vanderplas, A.~Passos,
  D.~Cournapeau, M.~Brucher, M.~Perrot, and E.~Duchesnay.
\newblock Scikit-learn: Machine learning in {P}ython.
\newblock \emph{Journal of Machine Learning Research}, 12:\penalty0 2825--2830,
  2011.

\bibitem[Rafailov et~al.(2023)Rafailov, Sharma, Mitchell, Ermon, Manning, and
  Finn]{rafailov2023direct}
Rafael Rafailov, Archit Sharma, Eric Mitchell, Stefano Ermon, Christopher~D
  Manning, and Chelsea Finn.
\newblock Direct preference optimization: Your language model is secretly a
  reward model.
\newblock \emph{arXiv preprint arXiv:2305.18290}, 2023.

\bibitem[Schulman et~al.(2017)Schulman, Wolski, Dhariwal, Radford, and
  Klimov]{schulman2017proximal}
John Schulman, Filip Wolski, Prafulla Dhariwal, Alec Radford, and Oleg Klimov.
\newblock Proximal policy optimization algorithms.
\newblock \emph{arXiv preprint arXiv:1707.06347}, 2017.

\bibitem[Siththaranjan et~al.(2023)Siththaranjan, Laidlaw, and
  Hadfield-Menell]{siththaranjan2023distributional}
Anand Siththaranjan, Cassidy Laidlaw, and Dylan Hadfield-Menell.
\newblock Distributional preference learning: Understanding and accounting for
  hidden context in rlhf.
\newblock \emph{arXiv preprint arXiv:2312.08358}, 2023.

\bibitem[Stiennon et~al.(2020)Stiennon, Ouyang, Wu, Ziegler, Lowe, Voss,
  Radford, Amodei, and Christiano]{stienon2020learning}
Nisan Stiennon, Long Ouyang, Jeff Wu, Daniel~M. Ziegler, Ryan Lowe, Chelsea
  Voss, Alec Radford, Dario Amodei, and Paul Christiano.
\newblock Learning to summarize from human feedback.
\newblock In \emph{NeurIPS}, 2020.

\bibitem[Tang et~al.(2024)Tang, Guo, Zheng, Calandriello, Munos, Rowland,
  Richemond, Valko, Pires, and Piot]{tang2024generalized}
Yunhao Tang, Zhaohan~Daniel Guo, Zeyu Zheng, Daniele Calandriello, R{\'e}mi
  Munos, Mark Rowland, Pierre~Harvey Richemond, Michal Valko,
  Bernardo~{\'A}vila Pires, and Bilal Piot.
\newblock Generalized preference optimization: A unified approach to offline
  alignment.
\newblock \emph{arXiv preprint arXiv:2402.05749}, 2024.

\bibitem[Team et~al.(2023)Team, Anil, Borgeaud, Wu, Alayrac, Yu, Soricut,
  Schalkwyk, Dai, Hauth, et~al.]{team2023gemini}
Gemini Team, Rohan Anil, Sebastian Borgeaud, Yonghui Wu, Jean-Baptiste Alayrac,
  Jiahui Yu, Radu Soricut, Johan Schalkwyk, Andrew~M Dai, Anja Hauth, et~al.
\newblock Gemini: a family of highly capable multimodal models.
\newblock \emph{arXiv preprint arXiv:2312.11805}, 2023.

\bibitem[Touvron et~al.(2023)Touvron, Martin, Stone, Albert, Almahairi, Babaei,
  Bashlykov, Batra, Bhargava, Bhosale, et~al.]{touvron2023llama}
Hugo Touvron, Louis Martin, Kevin Stone, Peter Albert, Amjad Almahairi, Yasmine
  Babaei, Nikolay Bashlykov, Soumya Batra, Prajjwal Bhargava, Shruti Bhosale,
  et~al.
\newblock Llama 2: Open foundation and fine-tuned chat models.
\newblock \emph{arXiv preprint arXiv:2307.09288}, 2023.

\bibitem[Wang et~al.(2024{\natexlab{a}})Wang, Xiong, Xie, Zhao, and
  Zhang]{wang2024armo}
Haoxiang Wang, Wei Xiong, Tengyang Xie, Han Zhao, and Tong Zhang.
\newblock Interpretable preferences via multi-objective reward modeling and
  mixture-of-experts.
\newblock \emph{arXiv preprint arXiv:2406.12845}, 2024{\natexlab{a}}.

\bibitem[Wang et~al.(2023)Wang, Dong, Zeng, Adams, Sreedhar, Egert, Delalleau,
  Scowcroft, Kant, Swope, and Kuchaiev]{wang2023helpsteer}
Zhilin Wang, Yi~Dong, Jiaqi Zeng, Virginia Adams, Makesh~Narsimhan Sreedhar,
  Daniel Egert, Olivier Delalleau, Jane~Polak Scowcroft, Neel Kant, Aidan
  Swope, and Oleksii Kuchaiev.
\newblock Helpsteer: Multi-attribute helpfulness dataset for steerlm, 2023.

\bibitem[Wang et~al.(2024{\natexlab{b}})Wang, Dong, Delalleau, Zeng, Shen,
  Egert, Zhang, Sreedhar, and Kuchaiev]{wang2024helpsteer2}
Zhilin Wang, Yi~Dong, Olivier Delalleau, Jiaqi Zeng, Gerald Shen, Daniel Egert,
  Jimmy~J. Zhang, Makesh~Narsimhan Sreedhar, and Oleksii Kuchaiev.
\newblock Helpsteer2: Open-source dataset for training top-performing reward
  models, 2024{\natexlab{b}}.

\bibitem[Weyssow et~al.(2024)Weyssow, Kamanda, and Sahraoui]{codeultrafeedback}
Martin Weyssow, Aton Kamanda, and Houari Sahraoui.
\newblock Codeultrafeedback: An llm-as-a-judge dataset for aligning large
  language models to coding preferences.
\newblock \emph{arXiv preprint arXiv:2403.09032}, 2024.

\bibitem[Zheng et~al.(2023)Zheng, Chiang, Sheng, Zhuang, Wu, Zhuang, Lin, Li,
  Li, Xing, Zhang, Gonzalez, and Stoica]{zheng2023judging}
Lianmin Zheng, Wei-Lin Chiang, Ying Sheng, Siyuan Zhuang, Zhanghao Wu, Yonghao
  Zhuang, Zi~Lin, Zhuohan Li, Dacheng Li, Eric Xing, Hao Zhang, Joseph~E.
  Gonzalez, and Ion Stoica.
\newblock Judging {LLM}-as-a-judge with {MT}-bench and chatbot arena.
\newblock In \emph{Thirty-seventh Conference on Neural Information Processing
  Systems Datasets and Benchmarks Track}, 2023.
\newblock URL \url{https://openreview.net/forum?id=uccHPGDlao}.

\bibitem[Ziegler et~al.(2019)Ziegler, Stiennon, Wu, Brown, Radford, Amodei,
  Christiano, and Irving]{ziegler2019fine}
Daniel~M Ziegler, Nisan Stiennon, Jeffrey Wu, Tom~B Brown, Alec Radford, Dario
  Amodei, Paul Christiano, and Geoffrey Irving.
\newblock Fine-tuning language models from human preferences.
\newblock \emph{arXiv preprint arXiv:1909.08593}, 2019.

\end{thebibliography}
\bibliographystyle{iclr2025_conference}

\newpage
\appendix

\section{Experimental Details}\label{supp:exp}

\paragraph{Attribute Training Datasets}
To train the quantile regression layers, we follow \cite{wang2024armo} and use the following training datasets with corresponding reward objectives.
\begin{itemize}
    \item \textbf{HelpSteer} \citep{wang2023helpsteer} (35k data): 
            helpsteer-helpfulness, helpsteer-correctness, helpsteer-coherence, helpsteer-complexity, helpsteer-verbosity
    \item \textbf{UltraFeedback} \citep{cui2023ultrafeedback} (240k data): 
        ultrafeedback-overall-score, ultrafeedback-instruction-following, ultrafeedback-truthfulness, ultrafeedback-honesty, ultrafeedback-helpfulness
    \item \textbf{BeaverTails-30k} \citep{ji2023beavertails} (30k data): 
        beavertails-is-safe
    \item \textbf{CodeUltraFeedback} \citep{codeultrafeedback} (50k data): 
        code-complexity, code-style, code-explanation, code-instruction-following, code-readability
    \item \textbf{Prometheus} \citep{prometheus} (200k data): 
        prometheus-score
    \item \textbf{Argilla-Capybara}\footnote{\url{https://hf.co/datasets/argilla/Capybara-Preferences-Filtered}} \citep{daniele2023amplify-instruct} (15k data): 
        argilla-overall-quality
    \item \textbf{Argilla-OpenOrca}\footnote{\url{https://hf.co/datasets/argilla/distilabel-intel-orca-dpo-pairs}} (13k data): 
    argilla-judge-lm
    \item \textbf{Argilla-Math-Preference}\footnote{\url{https://hf.co/datasets/argilla/distilabel-math-preference-dpo}} (2.4k data): This dataset shares the objective ultrafeedback-instruction-following with UltraFeedback
\end{itemize}
For each objective we normalize the attribute values in the range $[0,1]$.
Further, for each objective we limit the number of data points to $60,000$ for computational efficiency.

\paragraph{Training Data for the Gating Network} 
To train the gating network we again follow the \cite{wang2024armo} and train on the following datasets:
\begin{itemize}
    \item \textbf{HelpSteer} \citep{wang2023helpsteer} (37k pairs)
    \item \textbf{UltraFeedback} \citep{cui2023ultrafeedback} (340k pairs)
    \item \textbf{SHP} \citep{SHP} (93k pairs)
    \item \textbf{HH-RLHF} \citep{bai2022training,ganguli2022red} (157k pairs) 
    \item \textbf{PKU-SafeRLHF-30K} \citep{ji2023beavertails}
    \item \textbf{Argilla-Capybara} (15k pairs)
    \item \textbf{Argilla-Math-Preferences} (2.4k pairs)
    \item \textbf{CodeUltraFeedback} \citep{codeultrafeedback} (50k pairs)
    \item \textbf{PRM-Phase-2} \citep{lightman2023let} (80k pairs)
    \item \textbf{Prometheus2-Preference-Collection} \citep{kim2024prometheus} (200k pairs)
\end{itemize}

\end{document}